\documentclass{article}
\usepackage[a4paper,margin=1.2in]{geometry}

\usepackage{graphicx}
\usepackage{caption}
\usepackage{subcaption}
\usepackage{xspace}
\usepackage{algorithm}
\usepackage{algorithmicx}
\usepackage{algpseudocode}
\usepackage{microtype}
\usepackage{booktabs}
\usepackage{wrapfig}
\usepackage{natbib}

\usepackage{hyperref}

\title{Batch Stationary Distribution Estimation}

\author{
  Junfeng Wen$^{\dagger}$\thanks{Equal contribution.} \qquad 
  Bo Dai$^{\ddagger}$\footnotemark[1] \qquad 
  Lihong Li$^{\ddagger}$ \qquad 
  Dale Schuurmans$^{\dagger\ddagger}$\\
  $^\dagger$University of Alberta\qquad $^\ddagger$Google Research\\
  \texttt{junfengwen@gmail.com\ \{bodai,lihong,schuurmans\}@google.com}\\
}
\date{}

\usepackage{Definitions}
\usepackage{enumitem}

\usepackage[capitalise]{cleveref}

\usepackage{color}

\newcommand{\Dset}{\mathcal{D}}
\newcommand{\Xset}{\mathcal{X}}
\newcommand{\Sset}{\mathcal{S}}
\newcommand{\Aset}{\mathcal{A}}

\newcommand{\AlgName}{Variational Power Method\xspace}

\newcommand{\algshort}{VPM\xspace}

\renewcommand{\le}{\leqslant}
\renewcommand{\geq}{\geqslant}
\renewcommand{\ge}{\geqslant}

\begin{document}

\maketitle

\begin{abstract}

We consider the problem of approximating the stationary distribution of
an ergodic Markov chain given a set of sampled transitions. 
Classical simulation-based approaches assume access to the underlying process
so that trajectories of sufficient length can be gathered to
approximate stationary sampling.
Instead, we consider an alternative setting
where a \emph{fixed} set of transitions has been collected beforehand,
by a separate, possibly unknown procedure.
The goal is still to estimate properties of the stationary distribution,
but without additional access to the underlying system.
We propose a consistent estimator that is based on
recovering a correction ratio function over the given data.
In particular, we develop a variational power method (VPM)
that provides provably consistent estimates under general conditions.
In addition to unifying a number of existing approaches
from different subfields, we also find that VPM yields significantly better
estimates across a range of problems,
including queueing, stochastic differential equations, 
post-processing MCMC,
and off-policy evaluation.

\end{abstract}

\section{Introduction}\label{sec:intro}

Markov chains are a pervasive modeling tool in applied mathematics
of particular importance in stochastic modeling and machine learning.
A key property of an \emph{ergodic} Markov chain is the existence of a unique \emph{stationary distribution};
\ie, the long-run distribution of states that remains invariant
under the transition kernel.
In this paper,
we consider a less well studied but still important version of the 
stationary distribution estimation problem,
where one has access to a set of sampled transitions from a given Markov chain,
but does not know the mechanism by which the probe points were chosen,
nor is able to gather additional data from the underlying process.
Nevertheless, one would still like
to estimate target properties of the stationary distribution,
such as the expected value of a random variable of interest.

This setting is inspired by many practical scenarios where sampling from 
the Markov process is costly or unavailable,
but data has already been collected and available for analysis. 
A simple example is a queueing system consisting of a service desk that
serves customers in a queue.
Queue length changes stochastically as customers arrive or
leave after being served.
The long-term distribution of queue length
(\ie, the stationary distribution of the underlying Markov chain)
is the object of central interest for managing such a service~\citep{haviv2009queues,serfozo2009basics}. 
In practice, however, queue lengths are physical quantities that can only be
measured for moderate periods, perhaps on separate occasions,
but rarely for sufficient time to ensure the (stochastic)
queue length has reached the stationary distribution.
Since the measurement process itself is expensive,
it is essential to make reasonable inferences about
the stationary distribution from the collected data alone.

We investigate methods for estimating properties of the
stationary distribution solely from a batch of previously collected data.
The key idea is to first estimate a correction ratio function
over the given data,
which can then be used to estimate expectations of interest with
respect to the stationary distribution.
To illustrate, consider an ergodic Markov chain
with state space $\Xset$, transition kernel $\Tcal$,
and a unique stationary distribution $\mu$ that satisfies 
\begin{equation}\label{eq:stationary_dist}
\mu\rbr{x'} = \int \Tcal\rbr{x'|x}\mu\rbr{x}dx \defeq \rbr{\Tcal \mu}\rbr{x'}\,.
\end{equation}
Assume we are given a \emph{fixed} sample of state transitions, 
$\Dcal = \cbr{\rbr{x, x'}_{i=1}^n} \sim \Tcal\rbr{x'|x}p\rbr{x}$,
such that each $x$ has been sampled according to an \emph{unknown} probe
distribution $p$,
but each $x'$ has been sampled according to the true underlying
transition kernel, $x'|x\sim\Tcal\rbr{x'|x}$.
Below we investigate procedures for estimating the point-wise ratios,
$\widehat\tau\rbr{x_i}\approx\frac{\mu\rbr{x_i}}{p\rbr{x_i}}$,
such that the weighted empirical distribution
\[
\hat{\mu}(x) \defeq
\Big(\sum_{i=1}^n\widehat\tau\rbr{x_i}\Big)^{-1}
\sum_{i=1}^n \widehat\tau\rbr{x_i}\II\{x=x_i\}
\]
can be used to approximate $\mu$ directly,
or further used to estimate the expected value of some target function(s)
of $x$ with respect to $\mu$.
Crucially, the approach we propose
does not require knowledge of the probe distribution $p$,
nor does it require additional access to samples drawn from the
transition kernel $\Tcal$, 
yet we will be able to establish consistency
of the estimation strategy under general conditions.

In addition to developing the fundamental approach,
we demonstrate its applicability and efficacy in a range of
important scenarios beyond queueing, including:
\begin{itemize}[topsep=0pt,leftmargin=*, labelindent=0pt,itemindent=0pt]
\item {\bf Stochastic differential equations~(SDEs)}
SDEs are an essential modeling tool in many fields like
statistical physics~\citep{Kadanoff00},
finance~\citep{Oksendal13} and molecular dynamcis~\citep{Liu01}.
An autonomous SDE describes the instantaneous change of a random variable $X$ by
\begin{equation}\label{eq:wiener}
dX = f\rbr{X}dt + \sigma\rbr{X}dW\,,
\end{equation}
where $f\rbr{X}$ is a drift term, $\sigma\rbr{X}$ a diffusion term,
and $W$ the Wiener process.
Given data $\Dcal = \big\{\rbr{x, x'}_{i=1}^n\big\}$
such that $x\sim p\rbr{x}$ is drawn from an unknown probe distribution
and $x'$ is the next state after a small time step according to \eqref{eq:wiener},
we consider the problem of estimating quantities of
the stationary distribution $\mu$ when one exists.

\item {\bf Off-policy evaluation~(OPE)}
Another important application is \emph{behavior-agnostic off-policy evaluation}
\citep{NacChoDaiLi19} in reinforcement learning (RL).
Consider a Markov decision process (MDP)
specified by $M=\langle \Sset, \Aset, P, R \rangle$,
such that $\Sset$ and $\Aset$ are the state and action spaces,
$P$ is the transition function,
and $R$ is the reward function~\citep{Puterman14}.
Given a policy $\pi$ that maps $s\in\Sset$ to a
distribution over $\Aset$,
a random trajectory can be generated starting from an initial 
state $s_0$: $(s_0,a_0,r_0,s_1,a_1,r_1,\ldots)$,
where $a_t \sim \pi(\cdot|s_t)$, $s_{t+1} \sim
P(\cdot|s_t,a_t)$ and $r_t\sim R\rbr{s_t, a_t}$.  
The \emph{value} of a policy $\pi$ is defined to be its
long-term average per-step reward:
\[
\rho(\pi) \defeq \lim_{T\rightarrow\infty}\EE\sbr{\frac{1}{T}
\sum_{t=0}^{T-1} r_t} = \EE_{(s,a)\sim d_\pi\circ\pi}
\sbr{R(s,a)},
\]
where $d_\pi$ denotes the limiting distribution over
states $\Sset$ of the Markov process induced by $\pi$. 
In behavior-agnostic off-policy evaluation,
one is given a target policy $\pi$ and a set of transitions
$\Dset = \cbr{(s,a,r, s')_{i=1}^n}\sim
P\rbr{s'|s, a}p\rbr{s, a}$,
potentially generated by multiple behavior policies.   
From such data, an estimate for $\rho\rbr{\pi}$ can be formed
in terms of a stationary ratio estimator:
\begin{equation}\label{eq:avg_reward}
{\rho}(\pi) =
\EE_{\rbr{s, a}\sim p}\sbr{\frac{d_\pi\rbr{s}\pi\rbr{a|s}}
{p\rbr{s, a}}r\rbr{s, a}}
\approx\frac{1}{n} \sum_{i=1}^n \widehat{\tau}(s_i,a_i) r_i.
\end{equation}
We refer the interested readers to \secref{subsec:exp_ope} and 
\appref{appendix:ope}
for further discussion.
\end{itemize}

For the remainder of the paper,
we will outline four main contributions.
First, we generalize the classical power iteration method 
to obtain an algorithm, the \emph{\AlgName}~(\algshort),
that can work with arbitrary parametrizations in a functional space,
allowing for a flexible yet practical approach.
Second, we prove the consistency and convergence of~\algshort.
Third, we illustrate how a diverse set of
stationary distribution estimation problems, including those above,
can be addressed by \algshort in a unified manner.
Finally, we demonstrate empirically that \algshort
significantly improves estimation quality in a range of applications,
including queueing, sampling, SDEs and OPE.

\section{Variational Power Method}
\label{sec:power_ratio}

To develop our approach, first
recall the definition of $\Tcal$ and $\mu$ in \eqref{eq:stationary_dist}.
We make the following assumption about $\Tcal$ and $\mu$ throughout the paper.
\begin{assumption}[ergodicity]\label{asmp:markov_regularity}
The transition operator $\Tcal$ has a unique stationary distribution, denoted $\mu$.
\end{assumption}
Conditions under which this assumption holds are mild,
and have been extensively discussed in standard textbooks
\citep{meyn2009markov,LevPer17}.

Next, to understand the role of the probe distribution $p$,
note that we can always rewrite the stationary distribution
as $\mu=p\circ\tau$
(i.e., $\mu\rbr{x} \!=\! p\rbr{x}\tau\rbr{x}$,
hence $\tau\rbr{x} \!=\! \frac{\mu\rbr{x}}{p\rbr{x}}$),
provided the following assumption holds.
\begin{assumption}[absolute continuity]\label{asmp:base_regularity}
The stationary distribution $\mu$ is absolutely continuous w.r.t.\ $p$.
That is, there exists $C<\infty$ such that $\nbr{\tau}_\infty\le C$.
\end{assumption}
Assumption~\ref{asmp:base_regularity}
follows previous work~\citep{liu2017black,NacChoDaiLi19},
and is common in density ratio estimation 
\citep{sugiyama2008direct,gretton2009covariate} and 
off-policy evaluation \citep{wang2017optimal,xie2019towards}.

Combining these two assumptions,
definition~\eqref{eq:stationary_dist} yields
\begin{align}
\mu\rbr{x'} = \int \Tcal\rbr{x'|x}\mu\rbr{x}dx
&=\int \Tcal\rbr{x'|x}p\rbr{x}\frac{\mu\rbr{x}}{p\rbr{x}}dx
\defeq \int \Tcal_p\rbr{x, x'}\tau\rbr{x}dx,\nonumber
\\
\mbox{which implies}\quad
p\rbr{x'}\tau\rbr{x'} 
&= {\int\Tcal_p\rbr{x, x'}\tau\rbr{x}dx}\defeq \Tcal_p\tau \rbr{x'}.
\label{eq:stationary_ratio}
\end{align}
This development reveals how, under the two stated assumptions,
there is sufficient information
to determine the unique ratio function $\tau$ that ensures $p\circ\tau=\mu$
in principle.
Given such a function $\tau$,
we can then base inferences about $\mu$ solely
on data sampled from $p$ and $\tau$.

\subsection{Variational Power Iteration}
\label{subsec:algname}

To develop a practical algorithm for recovering $\tau$
from the constraint \eqref{eq:stationary_ratio}, in function space,
we first consider the classical power method
for recovering the $\mu$ that satisfies \eqref{eq:stationary_dist}.
From \eqref{eq:stationary_dist} it can be seen that the stationary
distribution $\mu$ is an eigenfunction of $\Tcal$.
Moreover, it is the \emph{principal} eigenfunction,
corresponding to the largest eigenvalue $\lambda_1=1$.
In the simpler case of finite $\Xset$,
the vector $\mu$ is the principal (right) eigenvector of the transposed transition matrix. 
A standard approach to computing $\mu$ is then the power method:
\begin{eqnarray}\label{eq:power_iteration}
\mu_{t+1} = \Tcal\mu_t
,
\end{eqnarray}
whose iterates converge to $\mu$ at a rate linear in $\abr{\lambda_2}$,
where $\lambda_2$ is the second largest eigenvalue of $\Tcal$.
For ergodic Markov chains, one has $|\lambda_2|<1$
\citep[Chap 20]{meyn2009markov}. 

Our initial aim is to extend this power iteration approach
to the constraint \eqref{eq:stationary_ratio}
without restricting the domain $\Xcal$ to be finite.
This can be naturally achieved by the update
\begin{equation}
\label{eq:power_ratio}
\tau_{t+1} = \frac{\Tcal_p\tau_t}{p},
\end{equation}
where the division is element-wise.
Clearly the fixed point of \eqref{eq:power_ratio}
corresponds to the solution of \eqref{eq:stationary_ratio}
under the two assumptions stated above.
Furthermore, just as for $\mu_t$ in \eqref{eq:power_iteration},
$\tau_t$ in \eqref{eq:power_ratio} also converges to $\tau$ at a linear rate
for finite $\Xcal$. 
Unfortunately,
the update~\eqref{eq:power_ratio} cannot be used directly
in a practical algorithm for two important reasons.
First, we do not have a point-wise evaluator for $\Tcal_p$,
but only samples from $\Tcal_p$.
Second, the operator $\Tcal_p$ is applied to a function $\tau_t$,
which typically involves an intractable integral over $\Xset$ in general.
To overcome these issues, we propose a variational method
that considers a series of reformulated problems whose optimal solutions
correspond to the updates~\eqref{eq:power_ratio}.

To begin to develop a practical variational approach,
first note that \eqref{eq:power_ratio} operates directly on the density ratio,
which implies the density ratio estimation techniques of
\citet{NguWaiJor08} and \citet{SugSuzKan12} can be applied.
Let $\phi$ be a lower semicontinuous, convex
function satisfying $\phi\rbr{1} = 0$,
and consider the induced $f$-divergence,
\begin{equation}
D_\phi\rbr{\tilde p\|\tilde q}
=
\int \tilde p\rbr{x}\phi\rbr{\frac{\tilde q\rbr{x}}{\tilde p\rbr{x}}}dx
=
-\rbr{\min_{\nu}\,\,\EE_{\tilde p}\sbr{\phi^*\rbr{\nu}} - \EE_{\tilde q}\sbr{\nu}},
\label{eq:f_divergence}
\end{equation}
where $\phi^*\rbr{x} = \sup_{y\in \RR} x^\top y - \phi\rbr{y}$
is the conjugate function of $\phi$.
The key property of this formulation is that 
for any disributions $\tilde p$ and $\tilde q$, 
the inner optimum in $\nu$ 
satisfies $\partial\phi^*(\nu)=\tilde q/\tilde p$ \citep{NguWaiJor08};
that is, the optimum in \eqref{eq:f_divergence}
can be used to directly recover the distribution ratio.

To apply this construction to our setting,
first consider solving a problem of the following form in the dual space:
\begin{align}
\nu_{t+1}
&=
{\displaystyle \arg\min_{\nu}}\ \EE_{p\rbr{x'}}\sbr{\phi^*\rbr{\nu\rbr{x'}}}
-
\EE_{\Tcal_p\rbr{x, x'}}\sbr{\partial\phi^*\rbr{\nu_t\rbr{x}}\cdot \nu\rbr{x'}}
\label{eq:general_ratio_power}
\\
&=
\ {\displaystyle \arg\min_{\nu}}\ \EE_{p\rbr{x'}}\sbr{\phi^*\rbr{\nu\rbr{x'}}}
-
\EE_{\Tcal_p\rbr{x, x'}\tau_t\rbr{x}}\sbr{\nu\rbr{x'}}
,
\label{eq:general_ratio_power_inductive}
\end{align}
where to achieve \eqref{eq:general_ratio_power_inductive} we have applied 
the inductive assumption that $\tau_t=\partial\phi^*(\nu_t)$.
Then, by the optimality property of $\nu_{t+1}$,
we know that the solution $\nu_{t+1}$ must satisfy
\begin{equation}
\partial\phi^*(\nu_{t+1})=\smallfrac{\Tcal_p\tau_t}{p}=\tau_{t+1}
\label{eq:dual_ratio_update}
,
\end{equation}
hence the updated ratio $\tau_{t+1}$ in \eqref{eq:power_ratio}
can be directly recovered from the dual solution $\nu_{t+1}$,
while also retaining the inductive property that
$\tau_{t+1}=\partial\phi^*(\nu_{t+1})$ for the next iteration.

These developments can be further simplified by considering the
specific choice $\phi^*\rbr{x} = x^2/2$,
which satisfies $\phi^*=\phi$ and
simplifies the overall update to
\begin{equation}
\label{eq:variational_ratio_power}
\tau_{t+1}=\arg\min_{\tau\ge 0}\ 
\smallfrac{1}{2}\EE_{p\rbr{x'}}\sbr{\tau^2\rbr{x'}}
- \EE_{\Tcal_p\rbr{x, x'}}
\sbr{\tau_t\rbr{x}\tau\rbr{x'}}.
\end{equation}
Crucially,
this variational update~\eqref{eq:variational_ratio_power}
determines the same update as \eqref{eq:power_ratio},
but overcomes the two aforementioned difficulties.
First,
it bypasses the direct evaluation of $\Tcal_p$ and $p$,
and allows these to be replaced by unbiased estimates of expectations
extracted from the data.
Second,
it similarly bypasses the intractability of the operator application
$\Tcal_p\tau_t$ in the functional space,
replacing this with an expectation of $\tau_t\circ\tau$ that can also be
directly estimated from the data.

We now discuss some practical refinements of the approach.

\subsection{Maintaining Normalization}\label{subsec:normalization}

One issue that we did not address is that the update~\eqref{eq:power_ratio}
is scale-invariant, and therefore so is the corresponding variational
update~\eqref{eq:variational_ratio_power}.
In particular,
if $\EE_p\sbr{\tau_0}=c$,
then $\EE_p\sbr{\tau_k}=c$ for all $k\geq1$.
To maintain normalization, it is natural to initialize $\tau_0$ with 
$\EE_p\sbr{\tau_0}=1$.
Unfortunately, the variational update cannot be solved exactly in general,
meaning that the scale can drift over iterations.
To address this issue, we explicitly ensure normalization
by considering a constrained optimization in place of
\eqref{eq:variational_ratio_power}.
\begin{equation}\label{eq:constrained_ratio}
\min_{\tau\ge0} \,\,\smallfrac{1}{2}\EE_{p\rbr{x'}}\sbr{\tau^2\rbr{x'}} 
- \EE_{\Tcal_p\rbr{x, x'}}\sbr{\tau_t\rbr{x}\tau\rbr{x'}} 
\quad
\st \quad \EE_{p\rbr{x}}\sbr{\tau\rbr{x}} = 1.
\end{equation}
Although it might appear that solving \eqref{eq:constrained_ratio}
requires one to solve a sequence of regularized problems
\begin{equation}\label{eq:regularized_ratio}
\min_{\tau\ge 0}\,\, \smallfrac{1}{2}\EE_{p\rbr{x'}}\sbr{\tau^2\rbr{x'}} 
- \EE_{\Tcal_p\rbr{x, x'}}\sbr{\tau_t\rbr{x}\tau\rbr{x'}}
+\lambda\rbr{\EE_p\sbr{\tau} - 1}^2
\end{equation}
with increasing $\lambda\rightarrow\infty$,
to ensure the constraint in \eqref{eq:constrained_ratio}
is satisfied exactly,
we note that this additional expense can be entirely avoided 
for the specific problem we are considering.

\begin{theorem}[Normalization of solution]\label{thm:consistency}
If $\EE_p\sbr{\tau_t}=1$,
then for any $\lambda >0 $,
the estimator~\eqref{eq:regularized_ratio} has the same
solution as~\eqref{eq:constrained_ratio},
hence  $\EE_p\sbr{\tau_{t+1}}=1$.
\end{theorem}

Hence, we can begin with any $\tau_0$ satisfying $\EE_p\sbr{\tau_0} = 1$
(e.g., $\tau_0 = \arg\min_{\tau} \rbr{\EE_p\sbr{\tau}- 1}^2$),
and the theorem ensures that the normalization of $\tau_{t+1}$
will be inductively maintained using any fixed $\lambda>0$. 
The proof is given in \appref{appendix:consistency}.

\subsection{Avoiding Double Sampling}\label{subsec:double_sampling}

Another issue is that, even though  the
problem \eqref{eq:constrained_ratio} is convex in $\tau$,
the penalty $\rbr{\EE_p\sbr{\tau}- 1}^2$
still presents a practical challenge,
since it involves a nonlinear function of an expectation.
In particular, its gradient
\[
\rbr{\EE_p\sbr{\tau} - 1}\EE_p\sbr{\nabla\tau}
\]
requires two \iid~samples from $p$ to obtain an unbiased estimate. 
To avoid the ``double sampling'' problem,
we exploit the fact that 
$x^2 = \max_{v\in \RR} 2xv - v^2$,
yielding the equivalent reformulation of \eqref{eq:regularized_ratio}:
\begin{equation}
\min_{\tau\ge 0}\max_{v\in \RR}\ J(\tau,v)
=\smallfrac{1}{2}\EE_{p\rbr{x'}}\sbr{\tau^2\rbr{x'}}
- \EE_{\Tcal_p\rbr{x, x'}}\sbr{\tau_t\rbr{x}\tau\rbr{x'}}
+ \lambda\sbr{2v\rbr{\EE_p\sbr{\tau} - 1} - v^2}.
\label{eq:saddle_ratio}
\end{equation}
Crucially, the dual variable $v$ is a scalar,
making this problem much simpler than dual embedding~\citep{dai2017learning},
where the dual variables form a parameterized
function that introduces approximation error.
The problem \eqref{eq:saddle_ratio} is a
straightforward convex-concave objective with respect to $\rbr{\tau, v}$
that can be optimized by stochastic gradient descent.

\subsection{Damped Iteration}\label{subsec:damped}

A final difficulty to be addressed
arises from the fact that, in practice,
we need to optimize the variational objective based on sampled data,
which induces approximation error
since we are replacing the true operator $\Tcal_p$ by a stochastic
estimate $\widehat{\Tcal}_p$ such that $\EE[\widehat{\Tcal}_p]=\Tcal_p$.
Without proper adjustment, such estimation errors can accumulate over the
power iterations, and lead to inaccurate results.

To control the error due to sampling,
we introduce a damped version of the update \citep{ryu2016primer},
where instead of performing a stochastic update
$\tau_{t+1}=\smallfrac{\widehat{\Tcal}_p}{p}\tau_{t}$,
we instead perform a damped update given by
\begin{equation}
\begin{split}
\tau_{t+1} 
&= (1-\alpha_{t+1})\cdot \tau_{t} 
+ \alpha_{t+1}\cdot \smallfrac{\widehat{\Tcal}_p}{p}\tau_{t}
\end{split}
\label{eq:damped_power_ratio}
\end{equation}
where $\alpha_t\in(0,1)$ is a stepsize parameter. 
Intuitively, the update error introduced by the stochasticity of
$\widehat{\Tcal}_p$ is now controlled by the stepsize $\alpha_t$. 
The choice of stepsize and convergence of the algorithm is discussed in
\secref{sec:convergence}.

The damped iteration can be conveniently implemented with minor modifications
to the previous objective. 
We only need to change the sample from $\Tcal_p$ in \eqref{eq:saddle_ratio} by a weighted sample:
\begin{equation}
\begin{split}
\min_{\tau\ge 0} \max_{v\in \RR}\ 
J(\tau,v)
&=
\smallfrac{1}{2}\EE_{p\rbr{x'}}\sbr{\tau^2\rbr{x'}}
- (1-\alpha_{t+1})\EE_{p\rbr{x'}}
\sbr{\tau_t\rbr{x'}\tau\rbr{x'}}
\\
&\quad- \alpha_{t+1}\EE_{\Tcal_p\rbr{x, x'}}
\sbr{\tau_t\rbr{x}\tau\rbr{x'}}
+ \lambda\sbr{2v\rbr{\EE_p\sbr{\tau} - 1} - v^2}
.
\end{split}
\label{eq:damped_saddle_ratio}
\end{equation}

\subsection{A Practical Algorithm}\label{subsec:prac_alg}

\begin{algorithm}[t]
\caption{Variational Power Method}\label{alg:VPM}
  \begin{algorithmic}[1]
  
  \State \textbf{Input}: Transition data $\Dcal=\{(x,x')_{i=1}^n\}$, learning rate $\alpha_\theta,\alpha_v$, number of power steps $T$, number of inner optimization steps $M$, batch size $B$

  \State Initialize $\tau_\theta$

  \For {$t=1 \dots T$}

    \State Update and fix the reference network $\tau_{t}=\tau_\theta$
    
    \For {$m=1 \dots M$}
    
      \State Sample transition data $\{(x,x')_{i=1}^B\}$
      
      \State Compute gradients $\nabla_\theta J$ and $\nabla_v J$ from \eqref{eq:tau_v_gradients}

      \State $\theta=\theta-\alpha_\theta\nabla_\theta J$
      \Comment{gradient descent}

      \State $v=v+\alpha_v\nabla_v J$
      \Comment{gradient ascent}
    
    \EndFor

  \EndFor
  \State Return $\tau_\theta$
  \end{algorithmic}
\end{algorithm}

A practical version of \algshort is described in \algref{alg:VPM}. 
It solves \eqref{eq:damped_saddle_ratio} using a parameterized
$\tau:\Xcal\mapsto\RR$
expressed as a neural network $\tau_\theta$ with parameters $\theta$.
Given the constraint $\tau\ge0$,
we added a softplus activation $\log(1+\exp(\cdot))$ to the final layer to ensure positivity.
The expectations with respect to $p$ and $\Tcal_p$ are
directly estimated from sampled data.
When optimizing $\tau_\theta$ by stochastic gradient methods,
we maintain a copy of the previous network $\tau_{t}$ as the reference network
to compute the third term of~\eqref{eq:damped_saddle_ratio}.
The gradients of $J(\tau,v)$ with respect to $\theta$ and
$v$ are given by
\begin{equation}
\label{eq:tau_v_gradients}
\begin{split}
\nabla_\theta J(\tau,v)
&=\EE_{p}\sbr{\tau\nabla_\theta\tau}
- (1-\alpha_{t+1})\EE_{p}\sbr{\tau_t \nabla_\theta \tau}
- \alpha_{t+1}\EE_{\Tcal_p}\sbr{\tau_t \nabla_\theta \tau}
+2\lambda v\EE_p\sbr{\nabla_\theta\tau},\\
\nabla_v J(\tau,v) 
&=2\lambda\rbr{\EE_p\sbr{\tau} - 1 - v}.
\end{split}
\end{equation}
After convergence of $\tau_\theta$ in each iteration,
the reference network is updated by setting $\tau_{t+1}=\tau_\theta$.
Note that one may apply other gradient-based optimizers instead of SGD.

\section{Convergence Analysis}
\label{sec:convergence}

We now demonstrate that the final algorithm obtains sufficient control
over error accumulation to achieve consistency.
For notation brevity, we discuss the
result for the simpler form \eqref{eq:power_iteration}
instead of the ratio form \eqref{eq:power_ratio}.
The argument easily extends to the ratio form.

Starting from the plain stochastic update
$\mu_t=\widehat{\Tcal}\mu_{t-1}$,
the damped update can be expressed by
\begin{equation}
\begin{split}
\mu_{t} 
&= (1-\alpha_t)\mu_{t-1} + \alpha_t \widehat{\Tcal}\mu_{t-1} \\
&= (1-\alpha_t)\mu_{t-1} + \alpha_t \Tcal\mu_{t-1} 
+ \alpha_t\epsilon
,
\end{split}
\label{eq:damped_power_iteration}
\end{equation}
where $\epsilon$ is the error due to stochasticity in $\widehat{\Tcal}$. 
The following theorem establishes the convergence properties of
the damped iteration.

\begin{theorem}[Informal]\label{thm:convergence} 
Under mild conditions,
after $t$ iteration with step-size $\alpha_t=1/\sqrt{t}$, we have
\begin{align*}
\EE \sbr{\nbr{\mu_R - \Tcal\mu_R}^2_2}
\le \frac{C_1}{\sqrt{t}}\nbr{\mu_0-\mu}_2^2 
+ \frac{C_2\ln t}{\sqrt{t}}\nbr{\epsilon}_2^2,
\end{align*}
for some constants $C_1,C_2>0$,
where the expectation is taken over the distribution of iterates
$(\mu_R)_{R=1}^t$.
In other words,
$\EE\sbr{\nbr{\mu_R - \Tcal\mu_R}^2_2} = \widetilde{\Ocal}\rbr{t^{-1/2}}$,
and consequently $\mu_R$ converges to $\mu$
for ergodic $\Tcal$ .
\end{theorem}

The precise version of the theorem statement, together with a complete proof,
is given in
\appref{appendix:convergence}.

Note that the optimization quality depends on the number of samples,
the approximation error of the parametric family,
and the optimization algorithm.
There is a complex trade-off between these factors~\citep{bottou2008tradeoffs}.
On one hand, with more data, the statistical error is reduced,
but the computational cost of the optimization increases.
On the other hand, with a more flexible parametrization, such as
neural networks, reduces the approximation error,
but adds to the difficulty of optimization 
as the problem might no longer be convex.
Alternatively,
if the complexity of the parameterized family is increased,
the consequences of statistical error also increases.

Representing $\tau$ in a reproducing kernel Hilbert space (RKHS)
is a particularly interesting case,
because the problem \eqref{eq:saddle_ratio} becomes convex,
hence the optimization error of the empirical surrogate is reduced to zero.
\citet[Theorem~2]{NguWaiJor08} show that, under mild conditions,
the statistical error can be bounded in rate
$\Ocal\rbr{n^{-\frac{1}{2+\beta}}}$ in terms of Hellinger distance
($\beta$ denotes the exponent in the bracket entropy of the RKHS),
while the approximation error will depend on the RKHS~\citep{Bach14}.

\section{Related Work}\label{subsec:related_work}

The algorithm we have developed
reduces distribution estimation to density ratio estimation, 
which has been extensively studied in numerous contexts.
One example is learning under covariate shift~\citep{Shimodaira00},
where the ratio $\tau$ can be estimated by different techniques~\citep{gretton2009covariate,NguWaiJor08,sugiyama2008direct,sugiyama2012machine}.
These previous works differ from the current setting
in that they require data to be sampled from both 
the target and proposal distributions.
By contrast, we consider a substantially more challenging problem,
where only data sampled from the proposal is available,
and the target distribution is given only \emph{implicitly}
by \eqref{eq:stationary_dist} through the transition kernel $\Tcal$. 
A more relevant approach is Stein importance sampling~\citep{liu2017black}, where the ratio is estimated by minimizing the kernelized Stein discrepancy~\citep{liu2016kernelized}. 
However, it requires additional gradient information about the target potential,
whereas our method only requires sampled transitions.
Moreover, the method of \citet{liu2017black}
is computationally expensive and does not extrapolate to new examples.

The algorithm we develop in this paper is inspired by the classic
power method for finding principal eigenvectors.
Many existing works have focused on the finite-dimension
setting~\citep{BalDasFre13, hardt2014noisy,YanBraZhaWan17}, 
while \citet{kim2005iterative} and \citet{XieLiaSon15}
have extended the power method
to the infinite-dimension case using RKHS.
Not only do these algorithms require access to the 
transition kernel $\Tcal$, but they also require tractable operator 
multiplications. 
In contrast, 
our method avoids direct interaction with the operator $\Tcal$,
and can use flexible parametrizations (such as neural networks)
to learn the density ratio without per-step renormalization.

Another important class of methods for estimating or sampling 
from stationary distributions are based on simulations.  A
prominent example is Markov chain Monte Carlo (MCMC),
which is widely used in many statistical inference 
scenarios~\citep{AndFreDouJor03,koller2009probabilistic,WelTeh11}.
Existing MCMC methods~\citep[\eg,][]{neal2011mcmc,hoffman2014no} require
repeated, and often many, interactions with the transition operator $\Tcal$ 
to acquire a single sample from the stationary distribution.
Instead, \algshort can be applied when only a fixed sample is available.
Interestingly, this suggests that \algshort can be used to 
``post-process'' samples generated from typical MCMC methods
to possibly make more effective use of the data. 
We demonstrated this possibility empirically in \secref{sec:exp}.
Unlike \algshort, other post-processing methods~\citep{oates2017control}
require additional information about the target distribution~\citep{RobCas04}.
Recent advances have also shown that learning parametric 
samplers can be beneficial~\citep{song2017nice,li2019adversarial},
but require the potential function.
In contrast, 
\algshort
directly learns the stationary
density ratio solely from transition data.

One important application of \algshort
is off-policy RL~\citep{precup2001off}.
In particular, in off-policy evaluation (OPE), one aims to evaluate a target
policy's performance, given data collected from a different behavior policy.
This problem matches our proposed framework
as the collected data naturally consists of 
transitions from a Markov chain, and one is interested in estimating quantities 
computed from the stationary distribution of a different policy.
(See \appref{appendix:ope} for a detailed description of how the VPM
algorithm can be applied to OPE, even when $\gamma=1$.)
Standard importance weighting is known to have high variance, 
and various techniques have been proposed 
to reduce variance~\citep{precup2001off,jiang2016doubly,
rubinstein2016simulation,thomas2016data,guo2017using}.
However, these methods still exhibit exponential variance
in the trajectory length~\citep{li2015toward,jiang2016doubly}.

More related to the present paper is the recent work on off-policy RL 
that avoids the exponential blowup of variance.
It is sufficient to adjust observed rewards according to the ratio 
between the target and behavior stationary
distributions~\citep{hallak2017consistent,liu2018breaking,gelada2019off}.
Unfortunately, these methods require knowledge of the behavior policy,
$p(a|s)$, in addition to the transition data,
which is not always available in practice.
In this paper, we focus on the behavior-agnostic scenario where $p(a|s)$
is unknown.
Although the recent work of \citet{NacChoDaiLi19} considers the same scenario,
their approach is only applicable when the discount factor $\gamma < 1$, 
whereas the method in this paper can handle any $\gamma\in[0,1]$.

\section{Experimental Evaluation}
\label{sec:exp}
\newcommand{\figwidth}{0.48\textwidth}
\newcommand{\onethirdfigwidth}{0.3\textwidth}
\newcommand{\onefourthfigwidth}{0.24\textwidth}
\newcommand{\onecolfourfigwidth}{0.116\textwidth}

In this section, we demonstrate the advantages of~\algshort in four representative applications.
Due to space limit, experiment details are provided in 
\appref{appendix:experiments}.

\subsection{Queueing}\label{subsec:exp_queue}

In this subsection, we use~\algshort to estimate the stationary distribution of queue length.
Following the standard Kendall's notation in queueing theory~\citep{haviv2009queues,serfozo2009basics}, 
we analyze the discrete-time Geo/Geo/1 queue, which is commonly used in the literature~\citep{atencia2004discrete,li2008analysis,wang2014equilibrium}.
Here the customer inter-arrival time and service time are geometrically distributed with one service desk.
The probe distribution $p(x)$ is a uniform distribution over the states in a predefined range $[0,B)$.
The observed transition $(x,x')$ is the length change in one time step. 
The queue has a closed-form stationary distribution that we can compare to~\citep[Sec.1.11]{serfozo2009basics}.

\begin{figure}[t]
\centering
\begin{subfigure}[b]{0.48\columnwidth}
\centering
\includegraphics[width=\textwidth]{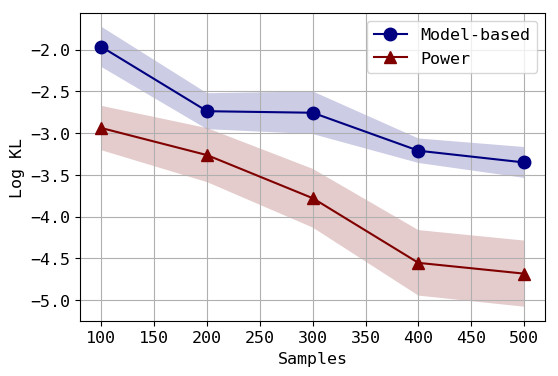}
\caption{Number of samples}
\label{subfig:queue_len}
\end{subfigure}
\begin{subfigure}[b]{0.48\columnwidth}
\centering
\includegraphics[width=\textwidth]{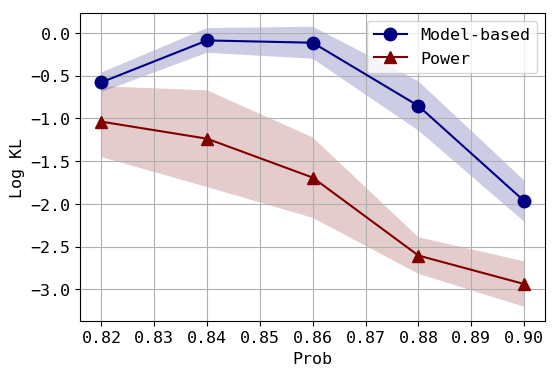}
\caption{Finish probability}
\label{subfig:queue_prob}
\end{subfigure}
\caption{Log KL divergence between estimation and the truth.}
\label{fig:queue}
\end{figure}

\cref{fig:queue} provides the log KL divergence between the estimated and true stationary distributions. 
We compare~\algshort to a model-based approach, which estimates the transition matrix $\widehat{\Tcal}(x'|x)$ 
from the same set of data, then simulates a long trajectory using $\widehat{\Tcal}$.
It can be seen that our method can be more effective across different sample sizes and queue configurations.

\subsection{Solving SDEs}\label{subsec:exp_sde}

We next apply VPM to solve a class of SDEs known as the Ornstein-Uhlenbeck process (OUP), which finds many applications in biology~\citep{butler2004phylogenetic}, financial mathematics and physical sciences~\citep{Oksendal13}. 
The process is described by the equation:
\[
dX=\theta(\mu-X)dt+\sigma dW
\]
where $\mu$ is the asymptotic mean, $\sigma>0$ is the deviation, $\theta>0$ determines the strength, and $W$ is the Wiener process.
The OUP has a closed-form solution,
which converges to the stationary distribution, a normal distribution $\Ncal(\mu,\sigma^2/2\theta)$, as $t\to\infty$. 
This allows us to conveniently calculate the Maximum Mean Discrepancy (MMD) between the adjusted sample to a true sample. 
We compare our method with the Euler-Maruyama (EM) method~\citep{gardiner2009stochastic}, which is a standard simulation-based method for solving SDEs.
\algshort uses samples from the EM steps to train the ratio network and the learned ratio is used to compute weighted MMD.

\begin{figure*}[t]
\centering
\begin{subfigure}[b]{\onefourthfigwidth}
\centering
\includegraphics[width=\textwidth]{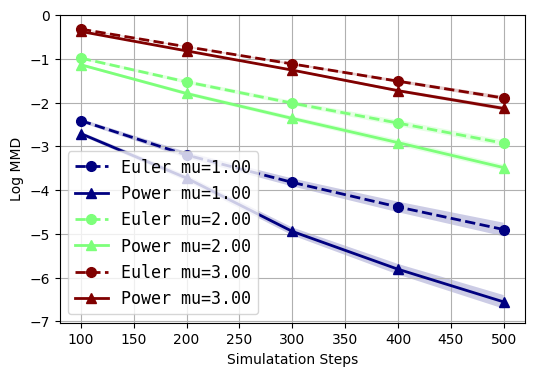}
\caption{Mean $\mu$}
\label{subfig:OUP_mu}
\end{subfigure}
\hfill
\begin{subfigure}[b]{\onefourthfigwidth}
\centering
\includegraphics[width=\textwidth]{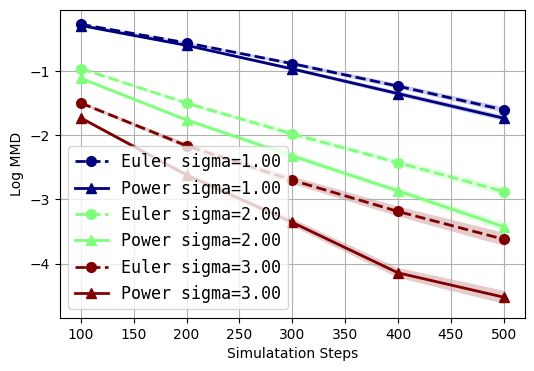}
\caption{Deviation $\sigma$}
\label{subfig:OUP_sigma}
\end{subfigure}
\hfill
\begin{subfigure}[b]{\onefourthfigwidth}
\centering
\includegraphics[width=\textwidth]{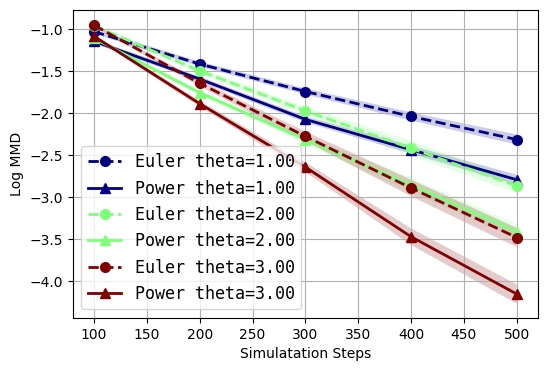}
\caption{Strength $\theta$}
\label{subfig:OUP_theta}
\end{subfigure}
\hfill
\begin{subfigure}[b]{\onefourthfigwidth}
\centering
\includegraphics[width=\textwidth]{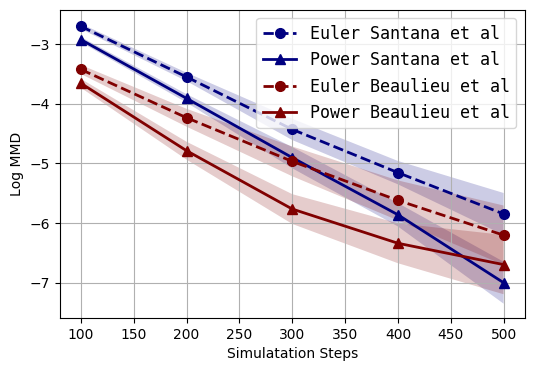}
\caption{Phylogeny Studies}
\label{subfig:OUP_bio}
\end{subfigure}
\caption{Log MMD versus number of EM steps across different settings, 
default $(\mu,\sigma,\theta)=(2,2,2)$. 
(d) is based on the real-world phylogeny studies~\citep{beaulieu2012modeling,santana2012dietary} 
with $(\mu,\sigma,\theta)=(0.618,1.584,3.85),(0.661, 0.710, 8.837)$ respectively.}
\label{fig:OUP}
\vspace{-2mm}
\end{figure*}

The results are shown in \cref{fig:OUP}, with different configurations of parameters $(\mu,\sigma,\theta)$.
It can be seen that \algshort consistently improves over the EM method in terms of the log MMD to a true sample from the normal distribution.
The EM method only uses the most recent data, which can be wasteful since the past data can carry additional information about the system dynamics.

In addition, we perform experiment on real-world phylogeny studies.
OUP is widely used to model the evolution of various organism traits. 
The results of two configurations~\citep[Tab.3\&1 resp.]{beaulieu2012modeling,santana2012dietary} are shown in \cref{subfig:OUP_bio}. 
Notably \algshort can improve over the EM method by correcting the sample with learned ratio.

\subsection{Post-processing MCMC}\label{subsec:exp_mcmc}

\begin{figure}[t]
\centering
\begin{subfigure}[b]{\onefourthfigwidth}
\centering
\includegraphics[width=\textwidth]{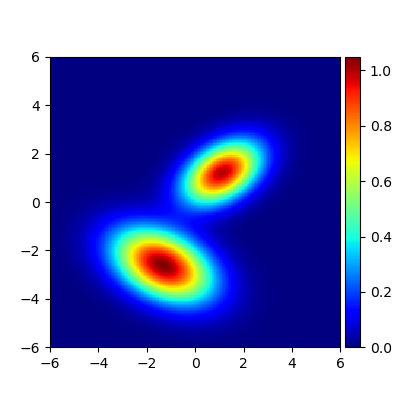}
\end{subfigure}
\begin{subfigure}[b]{\onefourthfigwidth}
\centering
\includegraphics[width=\textwidth]{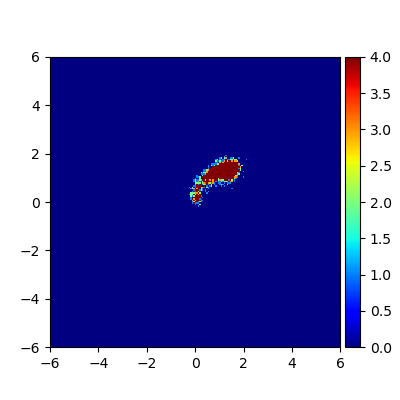}
\end{subfigure}
\begin{subfigure}[b]{\onefourthfigwidth}
\centering
\includegraphics[width=\textwidth]{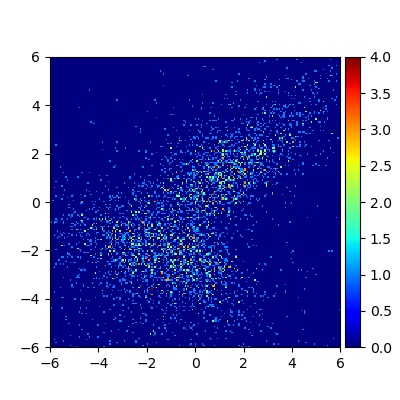}
\end{subfigure}
\begin{subfigure}[b]{\onefourthfigwidth}
\centering
\includegraphics[width=\textwidth]{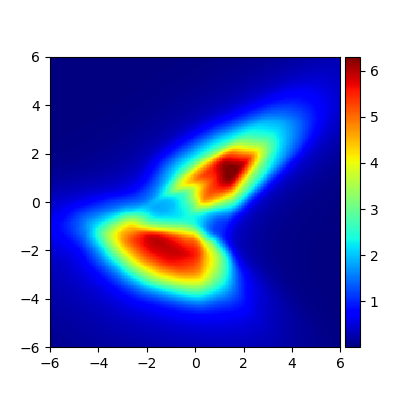}
\end{subfigure}
\begin{subfigure}[b]{\onefourthfigwidth}
\centering
\includegraphics[width=\textwidth]{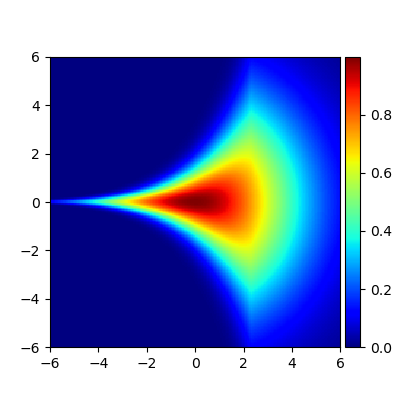}
\end{subfigure}
\begin{subfigure}[b]{\onefourthfigwidth}
\centering
\includegraphics[width=\textwidth]{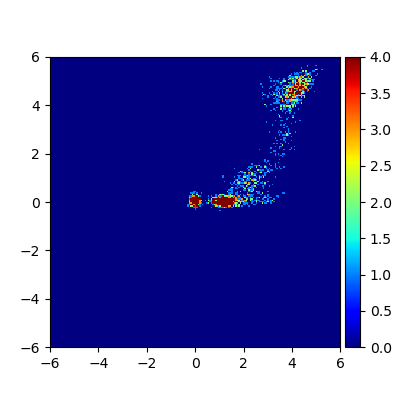}
\end{subfigure}
\begin{subfigure}[b]{\onefourthfigwidth}
\centering
\includegraphics[width=\textwidth]{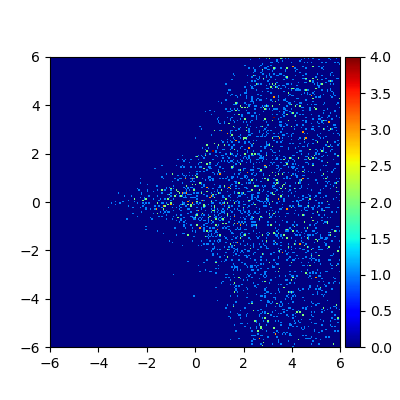}
\end{subfigure}
\begin{subfigure}[b]{\onefourthfigwidth}
\centering
\includegraphics[width=\textwidth]{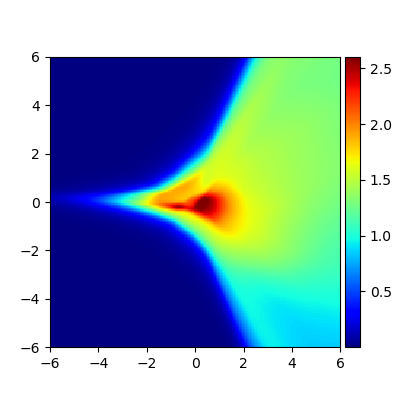}
\end{subfigure}
\begin{subfigure}[b]{\onefourthfigwidth}
\centering
\includegraphics[width=\textwidth]{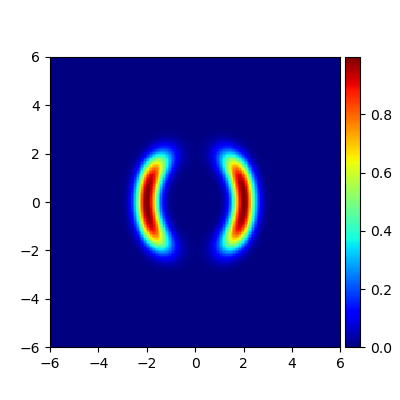}
\end{subfigure}
\begin{subfigure}[b]{\onefourthfigwidth}
\centering
\includegraphics[width=\textwidth]{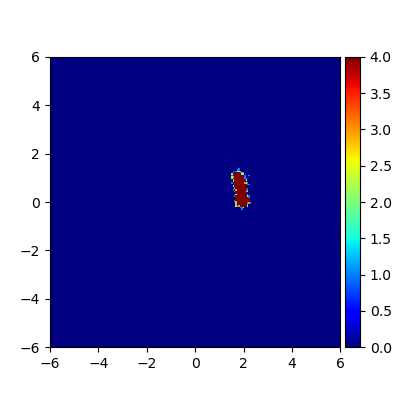}
\end{subfigure}
\begin{subfigure}[b]{\onefourthfigwidth}
\centering
\includegraphics[width=\textwidth]{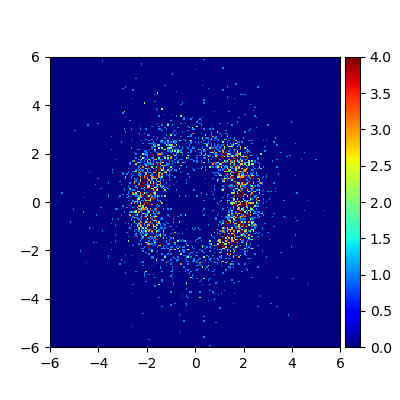}
\end{subfigure}
\begin{subfigure}[b]{\onefourthfigwidth}
\centering
\includegraphics[width=\textwidth]{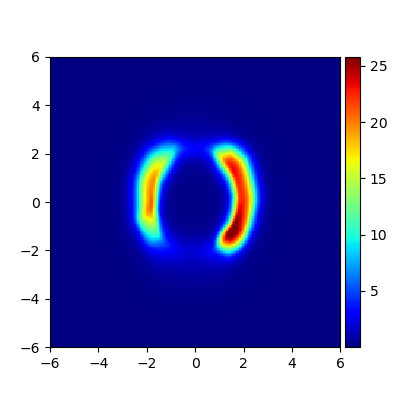}
\end{subfigure}
\begin{subfigure}[b]{\onefourthfigwidth}
\centering
\includegraphics[width=\textwidth]{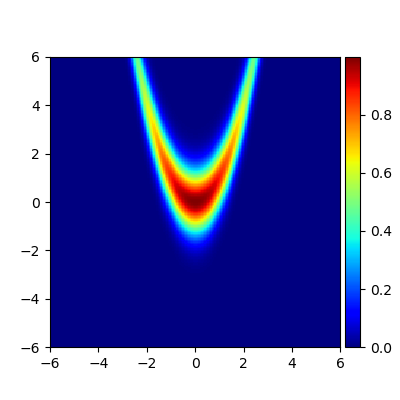}
\caption{Potentials}
\end{subfigure}
\begin{subfigure}[b]{\onefourthfigwidth}
\centering
\includegraphics[width=\textwidth]{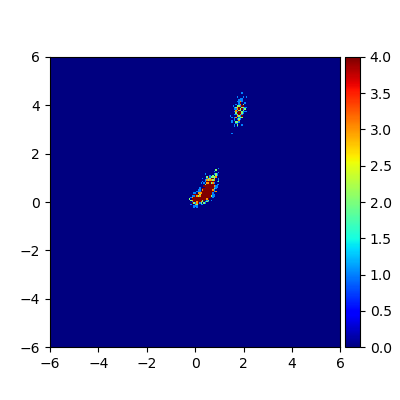}
\caption{Model}
\end{subfigure}
\begin{subfigure}[b]{\onefourthfigwidth}
\centering
\includegraphics[width=\textwidth]{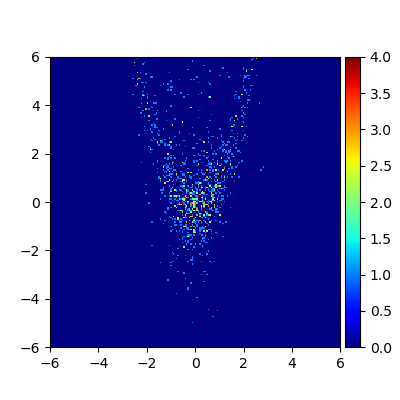}
\caption{\algshort}
\end{subfigure}
\begin{subfigure}[b]{\onefourthfigwidth}
\centering
\includegraphics[width=\textwidth]{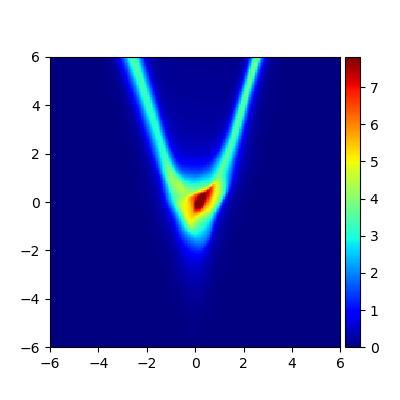}
\caption{Estimated $\tau$}
\end{subfigure}
\caption{The 2nd and 3rd columns are samples from the model-based method and \algshort respectively. Rows (from top to bottom) correspond to data sets: \textsf{2gauss, funnel, kidney, banana}.}
\label{fig:toy_mcmc}
\end{figure}

In this experiment, we demonstrate how \algshort can post-process MCMC to use transition data more effectively in order to learn the target distributions. 
We use four common potential functions as shown in the first column of \cref{fig:toy_mcmc}~\citep{neal2003slice,RezMoh15,LiSutStrGre18}.
A point is sampled from the uniform distribution $p(x)=\text{Unif}(x;[-6,6]^2)$, 
then transitioned through an HMC operator~\citep{neal2011mcmc}.
The transitioned pairs are used as training set $\Dset$.

We compare \algshort to a model-based method that explicitly learns a transition model $\widehat{\Tcal}(x'|x)$, parametrized as a neural network to produce Gaussian mean (with fixed standard deviation of $0.1$).
Then, we apply $\widehat{\Tcal}$ to a hold-out set drawn from $p(x)$ sufficiently many times, and use the final instances as limiting samples (second column of \cref{fig:toy_mcmc}). 
As for \algshort, since $p$ is uniform, the estimated $\widehat{\tau}$ is proportional to the true stationary distribution.
To obtain limiting samples (third column of \cref{fig:toy_mcmc}), we resample from a hold-out set drawn from $p(x)$ with probability proportional to $\widehat{\tau}$.

\begin{wrapfigure}{r}{0.45\textwidth}
\centering
\includegraphics[width=0.45\textwidth]{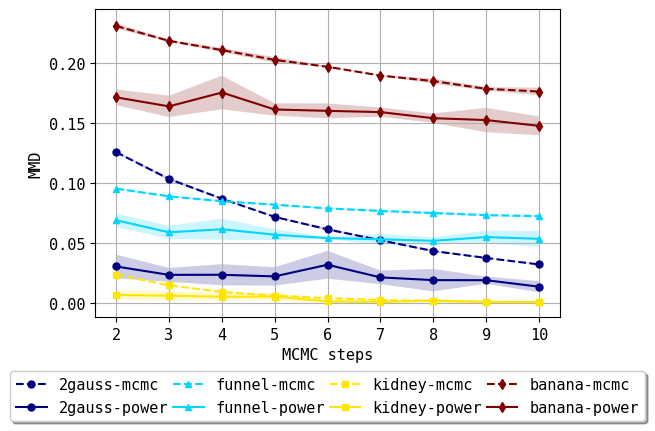}
\caption{MMD before and after ratio correction using \algshort.}
\label{fig:mcmc_mmd}
\end{wrapfigure}

The results are shown in \cref{fig:toy_mcmc}. 
Note that the model-based method quickly collapses all training data into high-probability regions as stationary distributions, which is an inevitable tendency of restricted parametrized $\widehat{\Tcal}$. 
Our learned ratio faithfully reconstructs the target density as shown in the right-most column of \cref{fig:toy_mcmc}.
The resampled data of \algshort are much more accurate and diverse than that of the model-based method.
These experiments show that \algshort can indeed effectively use a fixed set of data to recover the stationary distribution without additional information.


To compare the results quantitatively, \cref{fig:mcmc_mmd} shows the MMD of the estimated sample to a ``true'' sample. 
Since there is no easy way to sample from the potential function, the ``true'' sample consists of data after $2k$ HMC steps with rejection sampler.
After each MCMC step, \algshort takes the transition pairs as input and adjusts the sample importance according to the learned ratio. 
As we can see, after each MCMC step, \algshort is able to post-process the data and further reduce MMD by applying the ratio.
The improvement is consistent along different MCMC steps across different datasets.

\begin{figure}[ht]
\centering
\begin{subfigure}[b]{\onefourthfigwidth}
\centering
\includegraphics[width=\textwidth]{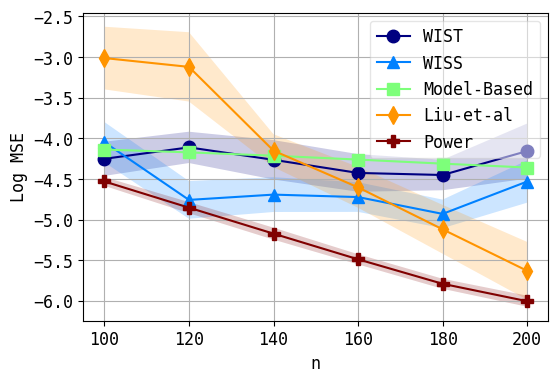}
\label{subfig:taxi_nt}
\end{subfigure}
\hfill
\begin{subfigure}[b]{\onefourthfigwidth}
\centering
\includegraphics[width=\textwidth]{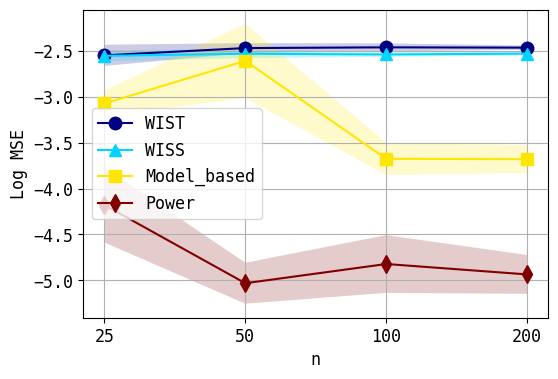}
\label{subfig:Reacher_n}
\end{subfigure}
\hfill
\begin{subfigure}[b]{\onefourthfigwidth}
\centering
\includegraphics[width=\textwidth]{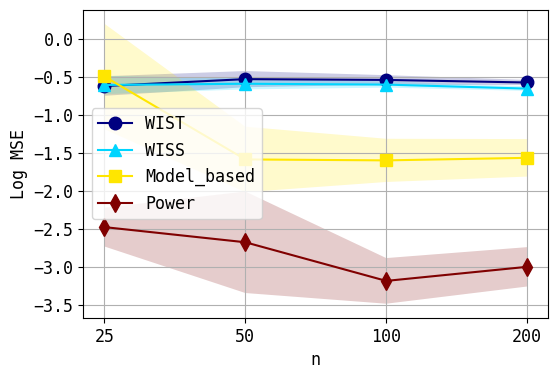}
\label{subfig:HalfCheetah_n}
\end{subfigure}
\hfill
\begin{subfigure}[b]{\onefourthfigwidth}
\centering
\includegraphics[width=\textwidth]{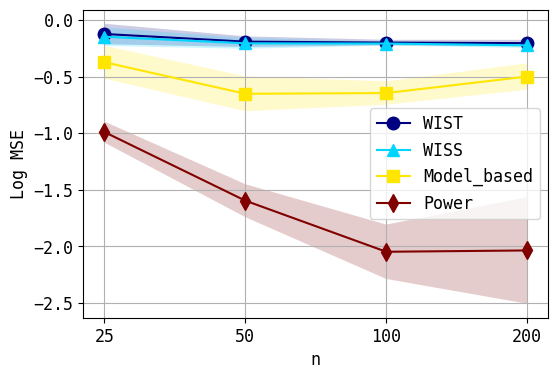}
\label{subfig:Ant_n}
\end{subfigure}
\begin{subfigure}[b]{\onefourthfigwidth}
\centering
\includegraphics[width=\textwidth]{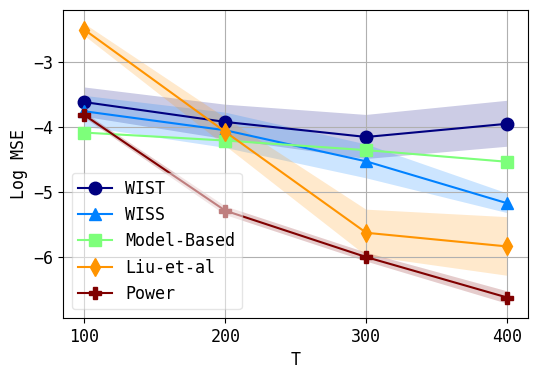}
\label{subfig:taxi_ts}
\end{subfigure}
\hfill
\begin{subfigure}[b]{\onefourthfigwidth}
\centering
\includegraphics[width=\textwidth]{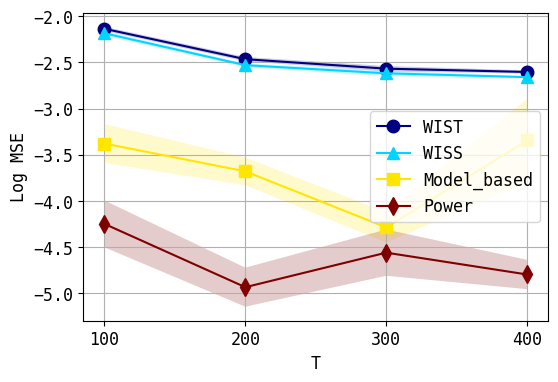}
\label{subfig:Reacher_T}
\end{subfigure}
\hfill
\begin{subfigure}[b]{\onefourthfigwidth}
\centering
\includegraphics[width=\textwidth]{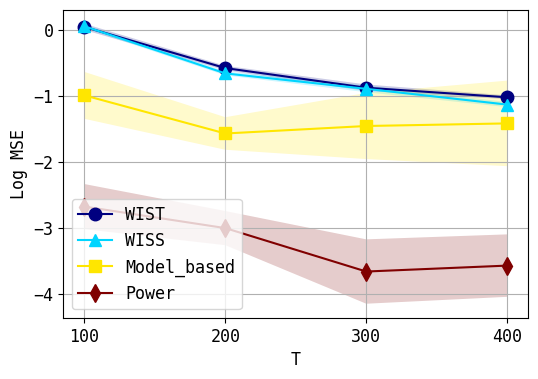}
\label{subfig:HalfCheetah_T}
\end{subfigure}
\hfill
\begin{subfigure}[b]{\onefourthfigwidth}
\centering
\includegraphics[width=\textwidth]{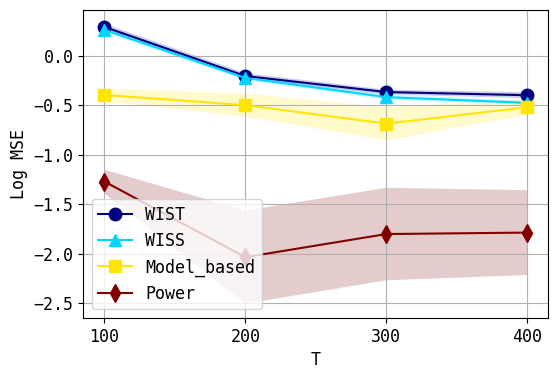}
\label{subfig:Ant_T}
\end{subfigure}
\begin{subfigure}[b]{\onefourthfigwidth}
\centering
\includegraphics[width=\textwidth]{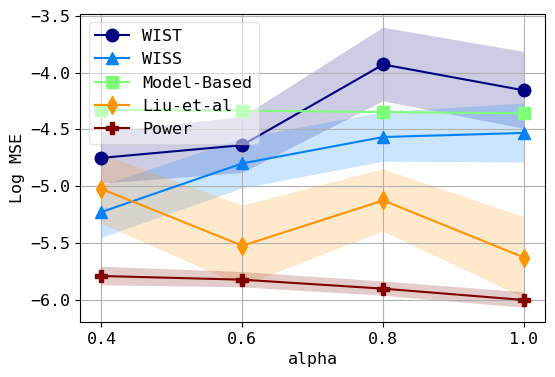}
\caption{Taxi}
\label{subfig:taxi_al}
\end{subfigure}
\hfill
\begin{subfigure}[b]{\onefourthfigwidth}
\centering
\includegraphics[width=\textwidth]{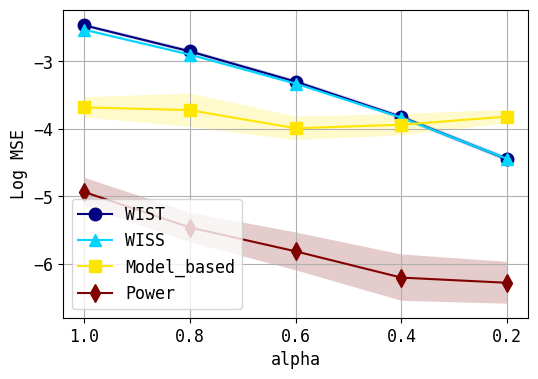}
\caption{Reacher}
\label{subfig:Reacher_al}
\end{subfigure}
\hfill
\begin{subfigure}[b]{\onefourthfigwidth}
\centering
\includegraphics[width=\textwidth]{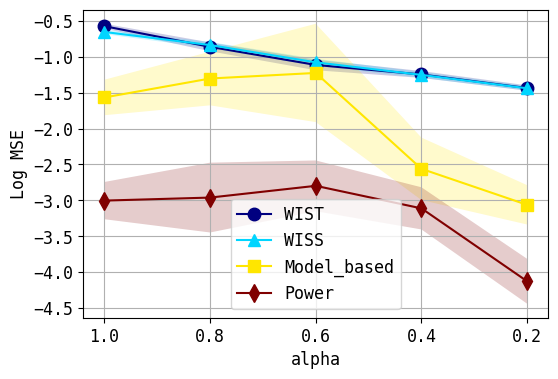}
\caption{HalfCheetah}
\label{subfig:HalfCheetah_al}
\end{subfigure}
\hfill
\begin{subfigure}[b]{\onefourthfigwidth}
\centering
\includegraphics[width=\textwidth]{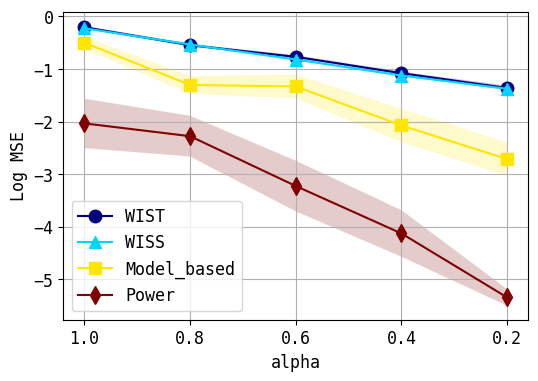}
\caption{Ant}
\label{subfig:Ant_al}
\end{subfigure}
\vspace{-2mm}
\caption{Log MSE of different methods for various datasets and settings.}
\label{fig:OPE}
\vspace{-2mm}
\end{figure}

\subsection{Off-Policy Evaluation}\label{subsec:exp_ope}

Finally, we apply our method to behavior-agnostic off-policy evaluation outlined in \secref{sec:intro}, in which only the transition data and the target policy are given, while the behavior policy is \emph{unknown}. 
Concretely, given a sample $\Dcal = \cbr{\rbr{s, a, r, s'}_{i=1}^n}$ from the behavior policy,
we compose each transition in $\Dcal$ with a target action $a'\sim \pi\rbr{\cdot|s'}$.
Denoting $x = \rbr{s, a}$, the data set can be expressed as
$\Dcal = \cbr{\rbr{x, x'}_{i=1}^n}$.
Applying the proposed~\algshort with $\Tcal(x'|x)$, we can estimate
$\frac{\mu\rbr{s, a}}{p\rbr{s, a}}$, hence
the average accumulated reward can be obtained via~\eqref{eq:avg_reward}.
Additional derivation and discussion can be found in 
\appref{appendix:ope}.

We conduct experiments on the (discrete) Taxi environment as in \citet{liu2018breaking}, and the challenging (continuous) environments including the Reacher, HalfCheetah and Ant.

Taxi is a gridworld environment in which the agent navigates to pick up and drop off passengers in specific locations.
The target and behavior policies are set as in \citet{liu2018breaking}.
For the continuous environments, the Reacher agent tries to reach a specified location by swinging an robotic arm, while the HalfCheetah/Ant agents are complex robots that try to move forward as much as possible.
The target policy is a pre-trained PPO or A2C neural network, which produces a Gaussian action distribution $\Ncal(m_t,\Sigma_t)$.
The behavior policy is the same as target policy but using a larger action variance $\Sigma_b=(1-\alpha)\Sigma_t+2\alpha\Sigma_t,\alpha\in(0,1]$.
We collect $T$ trajectories of $n$ steps each, using the behavior policy.

We compare \algshort to a model-based method that estimates both the transition $\Tcal$ and reward $R$ functions.
Using \emph{behavior cloning}, we also compare to the trajectory-wise and step-wise weighted importance sampling (WIST,WISS)~\citep{precup2001off}, as well as
\citet{liu2018breaking} with their public code for the Taxi environment.

The results are shown in \cref{fig:OPE}.
The $x$-axes are different configurations and the $y$-axes are the log Mean Square Error (MSE) to the true average target policy reward, estimated from abundant on-policy data collected from the target policy. 
As we can see, \algshort outperforms all baselines significantly across different settings, including number of trajectories, trajectory length and behavior policies. 
The method by \citet{liu2018breaking} can suffer from not knowing the behavior policy, as seen in the Taxi environment. 
Weighted importance sampling methods (WIST,WISS) also require access to the behavior policy.

\section{Conclusion}

We have formally considered the problem of 
estimating stationary distribution of an ergodic Markov chain
using a fixed set of transition data.
We extended a classical power iteration approach to the batch setting,
using an equivalent variational reformulation of the update rule to bypass
the agnosticity of transition operator and the intractable operations in
a functional space,
yielding a new algorithm \emph{\AlgName~(\algshort)}.
We characterized the convergence 
of~\algshort theoretically,
and demonstrated its empirical advantages for improving existing methods
on several important problems such as queueing, solving SDEs, post-processing MCMC and behavior-agnostic off-policy evaluation.

\newpage
\bibliographystyle{plainnat}
\bibliography{./ref,../../../../bibfile/bibfile}

\clearpage
\newpage
\appendix
\onecolumn
\begin{appendix}
\thispagestyle{plain}
\begin{center}
	{\huge Appendix}
\end{center}

\section{Consistency of the Objectives}\label{appendix:consistency}

\paragraph{\thmref{thm:consistency} (Consistency of solution)}
\textit{
If $\EE_p\sbr{\tau_t}=1$,
then for any $\lambda >0 $,
the estimator~\eqref{eq:regularized_ratio} has the same
solution as~\eqref{eq:constrained_ratio},
hence  $\EE_p\sbr{\tau_{t+1}}=1$.
}

\begin{proof}
Taking derivative of the objective function in \eqref{eq:constrained_ratio} and setting it to zero, we can see that the unconstrained solution is $\frac{\Tcal_p\tau_t}{p}$.
Moreover, it satisfies the constraint when  $\EE_p\sbr{\tau_t}=1$: we can rewrite $\tau_t = \frac{\mu_t}{p}$ for some distribution $\mu_t$ and $\EE_p\sbr{\frac{\Tcal_p\tau_t}{p}} = \int \Tcal\rbr{x'|x} \mu_t\rbr{x}dxdx' = 1$. 

We just need to show $\frac{\Tcal_p\tau_t}{p}$ is also the solution to~\eqref{eq:regularized_ratio}. Specifically, we have
\begin{equation}
\begin{split}
&\min_{\tau\ge 0}\,\, 
\frac{1}{2}\ 
\EE_{p\rbr{x'}}\sbr{\tau^2\rbr{x'}} 
- \EE_{\Tcal_p\rbr{x, x'}}\sbr{\tau_t\rbr{x}\tau\rbr{x'}}
+ \lambda\rbr{\EE_p\sbr{\tau} - 1}^2\\
&\ge \min_{\tau\ge 0}\,\, \frac{1}{2}\ 
\EE_{p\rbr{x'}}\sbr{\tau^2\rbr{x'}} 
- \EE_{\Tcal_p\rbr{x, x'}}\sbr{\tau_t\rbr{x}\tau\rbr{x'}}
+\min_{\tau\ge 0}\lambda\rbr{\EE_p\sbr{\tau} - 1}^2\\
&=\textstyle -\frac{1}{2}\EE_p\sbr{\rbr{\frac{\Tcal_p \tau_t}{p}}^2},
\end{split}
\label{eq:split_minimization}
\end{equation}
which can be attained by plugging in $\tau = \frac{\Tcal_p\tau_t}{p}$. 
Finally, we conclude the proof by noticing that~\eqref{eq:regularized_ratio} is strictly convex so the optimal solution is unique.
\end{proof}

\section{Convergence Analysis}\label{appendix:convergence}

Let $(X, \Sigma, \nu)$ be a measure space.
The $\Lcal^2(X)$ space consists of measurable functions $f:X\mapsto\RR$ such that $\|f\|=(\int|f|^2d\nu)^{1/2}<\infty$.
Suppose the initial $\mu_0\in\Lcal^2(X)$, we want to show the converging behavior of the following damped iteration:
\begin{equation}
\begin{split}
\mu_{t} 
&= (1-\alpha_t)\mu_{t-1} + \alpha_t \widehat{\Tcal}\mu_{t-1} \\
&= (1-\alpha_t)\mu_{t-1} + \alpha_t \Tcal\mu_{t-1} 
+ \alpha_t\epsilon
\end{split}
\label{eq:damped_update}
\end{equation}
with suitable step-sizes $\alpha_t\in(0,1)$, where $\epsilon\in\Lcal^2(X)$ is a random field due to stochacity in $\widehat{\Tcal}$. 
To this end, we will use the following lemma.

\begin{lemma}
\label{lemma:L2_expand}
For $\alpha\in\RR, f,g\in\Lcal^2(X)$
\[
\|(1-\alpha)f+\alpha g\|^2
=(1-\alpha)\|f\|^2+\alpha\|g\|^2
-\alpha(1-\alpha)\|f-g\|^2.
\]
\end{lemma}
\noindent
This can be proved by expanding both sides. Now we state our main convergence result.

\paragraph{\thmref{thm:convergence}}
\textit{
Suppose $\mu_0\in\Lcal^2(X)$, the step size is $\alpha_t=1/\sqrt{t}$, $\epsilon\in\Lcal^2(X)$ is a random field and $\Tcal$ has a unique stationary distribution $\mu$.
After $t$ iterations, define the probability distribution over the iterations as
\[
\Pr(R=k)=\frac{\alpha_k(1-\alpha_k)}{\sum_{k'=1}^t\alpha_{k'}(1-\alpha_{k'})}
\]
Then there exist some constants $C_1,C_2>0$ such that
\begin{align*}
\EE \sbr{\nbr{\mu_R - \Tcal\mu_R}^2_2}
\le \frac{C_1}{\sqrt{t}}\nbr{\mu_0-\mu}_2^2 
+ \frac{C_2\ln t}{\sqrt{t}}\nbr{\epsilon}_2^2,
\end{align*}
where the expectation is taken over $R$.
Consequently, $\mu_R$ converges to $\mu$ for ergodic $\Tcal$.
}

\begin{proof}
Using Lemma~\ref{lemma:L2_expand} and the fact that $\Tcal$ is non-expansive, we have
\[
\begin{split}
\|\mu_{t} - \mu\|^2
&=\|(1-\alpha_t)(\mu_{t-1}-\mu)
+\alpha_t(\Tcal\mu_{t-1}-\mu)
+\alpha_t\epsilon\|^2\\
&\le\|(1-\alpha_t)(\mu_{t-1}-\mu)
+\alpha_t(\Tcal\mu_{t-1}-\mu)\|^2
+\alpha_t^2\|\epsilon\|^2\\
&\le(1-\alpha_t)\|\mu_{t-1}-\mu\|^2
+\alpha_t\|\Tcal\mu_{t-1}-\mu\|^2
-\alpha_t(1-\alpha_t)\|\mu_{t-1}-\Tcal\mu_{t-1}\|^2
+\alpha^2_t\|\epsilon\|^2\\
&\le\|\mu_{t-1}-\mu\|^2
-\alpha_t(1-\alpha_t)\|\mu_{t-1}-\Tcal\mu_{t-1}\|^2
+\alpha^2_t\|\epsilon\|^2.
\end{split}
\]
Then telescoping sum gives
\[
0\le\|\mu_{t} - \mu\|^2
\le\|\mu_0 - \mu\|^2 + \sum_{k=1}^{t}\alpha_k^2\|\epsilon\|^2
-\sum_{k=1}^t\alpha_k(1-\alpha_k)\|\mu_k-\Tcal\mu_k\|^2
\]
So
\[
\begin{split}
\sum_{k=1}^t\alpha_k(1-\alpha_k)\|\mu_k-\Tcal\mu_k\|^2
&\le\|\mu_0 - \mu\|^2
+\sum_{k=1}^{t}\alpha_k^2\|\epsilon\|^2.
\end{split}
\]
Divide both sides by $\sum_{k=1}^t\alpha_k(1-\alpha_k)$ (taking expectation over iterations) gives
\[
\begin{split}
\EE[\|\mu_R-\Tcal\mu_R\|^2]
=\sum_{k=1}^t
\frac{\alpha_k(1-\alpha_k)}{\sum_{k'}\alpha_{k'}(1-\alpha_{k'})}
\|\mu_k-\Tcal\mu_k\|^2
&\le\frac{\|\mu_0 - \mu\|^2
+\sum_{k=1}^{t}\alpha_k^2\|\epsilon\|^2}{\sum_{k=1}^{t}\alpha_{k}(1-\alpha_{k})}.
\end{split}
\]
When $\alpha_t=1/\sqrt{t}$, we have 
\[
\begin{split}
\sum_{k=1}^{t}\alpha_k^2\|\epsilon\|^2
&=\sum_{k=1}^{t}\frac{1}{k}\|\epsilon\|^2
\le(\ln t + 1)\|\epsilon\|^2\\
\sum_{k=4}^{t}\alpha_k(1-\alpha_k)
&=\sum_{k=4}^{t}\frac{1}{\sqrt{k}}-\frac{1}{k}
\ge\int_4^t \left(\frac{1}{\sqrt{k+1}}-\frac{1}{k+1}\right) dk
=\Omega\left(t^{\frac{1}{2}}\right)
\end{split}
\]
So for big enough $t$, there exists $C_0>0$ such that
\[
\EE\sbr{\|\mu_R-\Tcal\mu_R\|^2}
\le \frac{\|\mu_0 - \mu\|^2
+\ln(t+1)\|\epsilon\|^2}{C_0\sqrt{t}},
\]
which leads to the the bound in the theorem and $\EE\sbr{\|\mu_R-\Tcal\mu_R\|^2}=\widetilde{\Ocal}\rbr{t^{-1/2}}$.
Additionally, since $\Tcal$ has a unique stationary distribution $\mu=\Tcal\mu$, we have $\mu_R$ converges to $\mu$.
\end{proof}

\section{Application to Off-policy Stationary Ratio Estimation}\label{appendix:ope}

We provide additional details describing how the variational power method
we have developed in the main body of the paper can be applied to the
behavior-agnostic off-policy estimation problem (OPE). 
The general framework has been introduced in \secref{sec:intro} and the implementation for the undiscounted case ($\gamma=1$) is demonstrated in \secref{subsec:exp_ope}. 
Specifically, given a sample $\Dcal = \cbr{\rbr{s, a, r, s'}_{i=1}^n}$ from the behavior policy,
we compose each transition in $\Dcal$ with a target action $a'\sim \pi\rbr{\cdot|s'}$.
Denoting $x = \rbr{s, a}$, the data set can be expressed as
$\Dcal = \cbr{\rbr{x, x'}_{i=1}^n}$.
Applying the proposed~\algshort with $\Tcal(x'|x)$, we can estimate
$\frac{\mu\rbr{s, a}}{p\rbr{s, a}}$. 
Here the $\mu(s,a)=d_\pi(s)\pi(a|s)$ consists of the stationary state occupancy $d_\pi$ and the target policy $\pi$, while $p(s,a)$ is the data-collecting distribution.
Then the average accumulated reward can be obtained via~\eqref{eq:avg_reward}.

Here we elaborate on how the discounted case (\ie, $\gamma\in(0,1)$) can be handled by our method.
We first introduce essential quantities 
similar to the undiscounted setting. 
For a trajectory generated stochastically
using policy $\pi$ from an initial state $s_0$:
$(s_0,a_0,r_0,s_1,a_1,r_1,\ldots)$,
where $a_t \sim \pi(\cdot|s_t)$, $s_{t+1} \sim P(\cdot|s_t,a_t)$
and $r_t\sim R\rbr{s_t, a_t}$,
the the policy value is
\[
\textstyle
\rho_\gamma\rbr{\pi}\defeq \rbr{1 - \gamma}
\EE_{s_0\sim\mu_0, a\sim\pi,s'\sim P}
\sbr{\sum_{t=0}^\infty \gamma^t r_t},
\]
where $\mu_0$ is the initial-state distribution.
Denote 
$$
d^\pi_t\rbr{s, a} = \PP\rbr{s_t=s, a_t=a\middle| 
\begin{bmatrix}
s_0\sim \mu_0, \forall i< t, \\
a_i\sim\pi\rbr{\cdot|s_i},\\
s_{i+1}\sim P(\cdot|s_i, a_i)
\end{bmatrix}
}.
$$ 
The discounted occupancy distribution is
\begin{equation}
\label{eq:mdp_stationary}
\mu_\gamma\rbr{s, a}\defeq (1-\gamma)\sum_{t=0}^{\infty}\gamma^t d^\pi_t\rbr{s, a}.
\end{equation}
Then, we can re-express the discounted accumulated reward 
via $\mu_\gamma$ and the stationary density ratio, 
\begin{equation}
\label{eq:discounted_ope}
\rho_\gamma\rbr{\pi} 
= \EE_{\rbr{s, a}\sim\mu_\gamma\rbr{s, a}}\sbr{r(s, a)}
= \EE_{\rbr{s, a}\sim p\rbr{s, a}}
\sbr{\frac{\mu_\gamma\rbr{s,a}}{p\rbr{s, a}}r(s, a)}.
\end{equation}

The proposed~\algshort is applicable to estimating the density ratio in this discounted case. 
Denoting $x = \rbr{s, a}$, $x' = \rbr{s', a'}$ respectively
for notational consistency, 
we expand $\mu_\gamma$ and use the definition of $d_t^\pi$:
\begin{align}\label{eq:discounsed_stationary_ratio}
\mu_\gamma\rbr{s', a'} 
&= \rbr{1 - \gamma}\mu_0\rbr{s'}\pi\rbr{a'|s'}
+\ \gamma \int \pi\rbr{a'|s'}P\rbr{s'|s, a}\mu_\gamma\rbr{s, a}ds\ da
\nonumber \\ 
\Longrightarrow p\rbr{x'}\tau^*\rbr{x'} 
&= \rbr{1 - \gamma}\mu_0\pi\rbr{x'}
+ \gamma\int \Tcal_{p}\rbr{x, x'}\tau^*\rbr{x}dx,
\end{align}
where $\mu_0\pi\rbr{x'} =\mu_0\rbr{s'}\pi\rbr{a'|s'}$
and $\Tcal_{p}\rbr{x, x'} = \pi\rbr{a'|s'}P\rbr{s'|s, a}p\rbr{s, a}$.

It has been shown that the RHS of~\eqref{eq:discounsed_stationary_ratio} is
contractive~\citep{SutBar98,MohRosTal12}, therefore, the fix-point iteration,
\begin{equation}\label{eq:discounted_power}
p\rbr{x'}\tau_{t+1}\rbr{x'}= \rbr{1 - \gamma}\mu_0\pi\rbr{x'}+ \gamma\int \Tcal_p\rbr{x, x'}\tau_t\rbr{x}dx,
\end{equation}
converges to the true $\tau$ as $t\rightarrow \infty$, provided the update above is carried out exactly. 
Compared to~\eqref{eq:power_ratio}, we can see that the RHS of~\eqref{eq:discounted_power} is now a mixture of $\mu_0\pi$ and $\Tcal_p$, with respective coefficients $(1-\gamma)$ and $\gamma$.

Similarly, we construct the $(t+1)$-step variational update as
\begin{equation}
\label{eq:discounted_variational_ratio_power}
\tau_{t+1} = \arg\min_{\tau\ge 0}\ 
\smallfrac{1}{2}\EE_{p\rbr{x'}} \sbr{\tau^2 \rbr{x'}}
- \gamma\EE_{\Tcal_{p}\rbr{x, x'}}
\sbr{\tau_t \rbr{x} \tau \rbr{x'}}
-\rbr{1 - \gamma}\EE_{\mu_0p\rbr{x'}}\sbr{\tau\rbr{x'}}
+\lambda\rbr{\EE_p\sbr{\tau} - 1}^2.
\end{equation}
Compared to~\eqref{eq:variational_ratio_power}, we see that the main difference
is the third term of~\eqref{eq:discounted_variational_ratio_power} involves the initial distribution.
As $\gamma\rightarrow 1$,~\eqref{eq:discounted_variational_ratio_power}
reduces to~\eqref{eq:variational_ratio_power}.

\section{Experiment Details}\label{appendix:experiments}

Here we provide additional details about the experiments. In all experiments, the regularization $\lambda=1$ and the optimizer is Adam with $\beta_1=0.5$. 

\subsection{Queueing}
For Geo/Geo/1 queue, when the arrival and finish probabilities are $q_a,q_f\in(0,1)$ respectively with $q_f>q_a$, the stationary distribution is $P(X=i)=(1-\rho)\rho^i$ where $\rho=q_a(1-q_f)/[q_f(1-q_a)]$~\citep[Sec.1.11]{serfozo2009basics}. 
The defaults are $(n,q_a,q_f)=(100,0.8,0.9)$ for the figures.
$\rho$ is called \emph{traffic intensity} in the queueing literature and we set $B=\lceil 40\rho\rceil$ in the experiment. The mean and standard error of the log KL divergence is computed based on 10 runs. 
We conduct closed-form update for $1000$ steps. 
As for the model-based method, we simulate the transition chain for $200$ steps to attain the estimated stationary distribution.

\subsection{Solving SDEs}
Using initial samples are uniformly spaced in $[0,1]$, we run the Euler-Maruyama (EM) method and evaluate the MMD along the path.
The $\tau$ model is a neural network with 2 hidden layers of 64 units each with ReLU and Softplus for the final layer. Numbers of outer and inner steps are $T=50,M=10$. The learning rate is 0.0005. At each evaluation time step $t$, we use the most recent $1\%$ of evolution data to train our model $\tau$. The plots are reporting the mean and standard deviation over 10 runs. For the phylogeny studies, the number of particles is $1k$ and $dt=0.0005$ for the EM simulation, while the rest settings using $dt=0.001$.

\subsection{Post-processing MCMC}
The potential functions are collected from several open-source projects\footnote{\url{https://github.com/kamenbliznashki/normalizing\_flows}}\footnote{\url{https://github.com/kevin-w-li/deep-kexpfam}}.
$50k$ examples are sampled from the uniform distribution $p(x)=\text{Unif}(x;[-6,6]^2)$, then transition each $x$ through an HMC operator (one leapfrog step of size $0.5$).
The $\tau$ model is a neural network with 4 hidden layers of 128 units each with ReLU activation and softplus activation for the output. The model-based $\widehat{\Tcal}$ has a similar structure except the final layer has 2D output without activation to estimate the Gaussian mean. The mini-batch size is $B=1k$, the maximum number of power iterations $T=150$ and the number of inner optimization steps is $M=10$. The model-based $\Tcal$ is given the same number of iterations ($MT=1500$). The learning rate is 0.001 for $\tau$ and 0.0005 for $\widehat{\Tcal}$. 
To compute the model-based sample, we apply the estimated transition $100$ time steps. The MMD plot is based on a ``true sample'' of size $2k$ from the stationary distribution (estimated by 2k HMC transition steps). The numbers are mean and standard deviation over 10 runs.
The MMD is computed by the Gaussian kernel with the median pairwise distance as kernel width. 

The quality of the transition kernel and the generated data is critical.
Since $x$ and $x'$ are supposed to be related, we use an HMC kernel with one leap-frog step.
The initial $x$ is effectively forgotten if using too many leap-frog steps. 
The main point is to show that our method can utilize the intermediate samples from the chain other than the final point. 
Moreover, to conform with Assumption 2, the potential functions are numerically truncated.

\begin{figure}[t]
\centering
\begin{subfigure}[b]{\onefourthfigwidth}
\centering
\includegraphics[width=\textwidth]{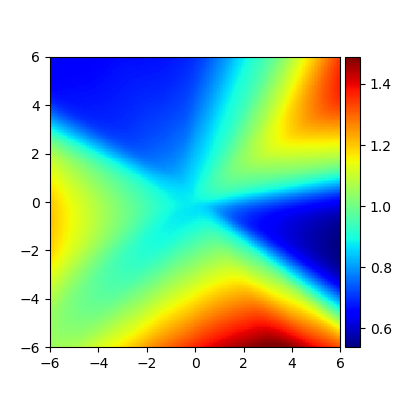}
\end{subfigure}
\begin{subfigure}[b]{\onefourthfigwidth}
\centering
\includegraphics[width=\textwidth]{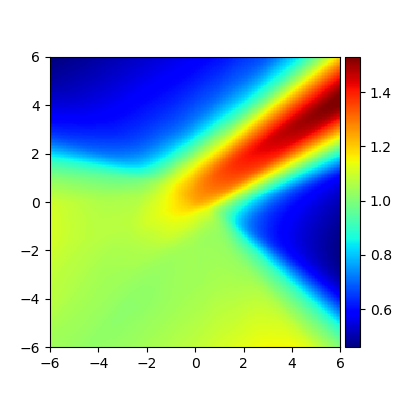}
\end{subfigure}
\begin{subfigure}[b]{\onefourthfigwidth}
\centering
\includegraphics[width=\textwidth]{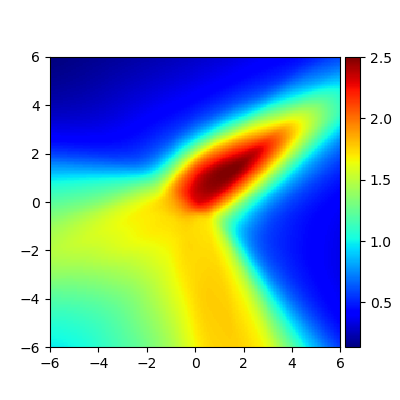}
\end{subfigure}
\begin{subfigure}[b]{\onefourthfigwidth}
\centering
\includegraphics[width=\textwidth]{figures/mcmc/2gauss_ratio_150.png}
\end{subfigure}
\begin{subfigure}[b]{\onefourthfigwidth}
\centering
\includegraphics[width=\textwidth]{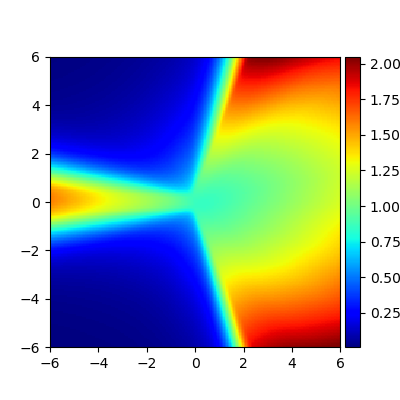}
\end{subfigure}
\begin{subfigure}[b]{\onefourthfigwidth}
\centering
\includegraphics[width=\textwidth]{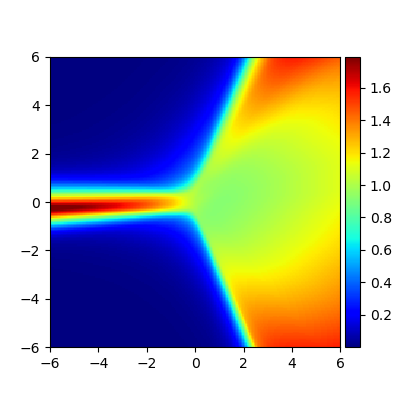}
\end{subfigure}
\begin{subfigure}[b]{\onefourthfigwidth}
\centering
\includegraphics[width=\textwidth]{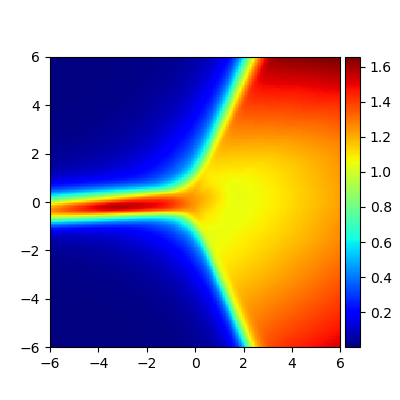}
\end{subfigure}
\begin{subfigure}[b]{\onefourthfigwidth}
\centering
\includegraphics[width=\textwidth]{figures/mcmc/funnel_ratio_150.png}
\end{subfigure}
\begin{subfigure}[b]{\onefourthfigwidth}
\centering
\includegraphics[width=\textwidth]{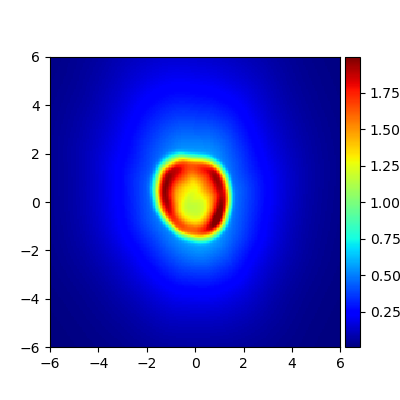}
\end{subfigure}
\begin{subfigure}[b]{\onefourthfigwidth}
\centering
\includegraphics[width=\textwidth]{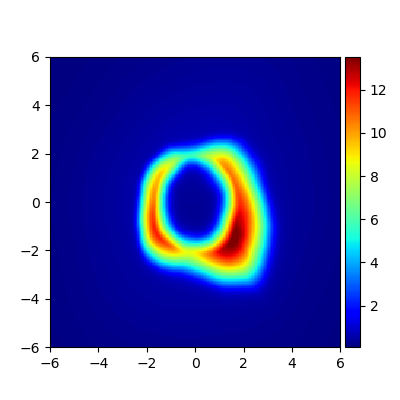}
\end{subfigure}
\begin{subfigure}[b]{\onefourthfigwidth}
\centering
\includegraphics[width=\textwidth]{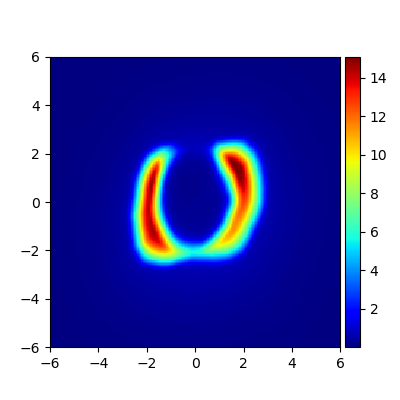}
\end{subfigure}
\begin{subfigure}[b]{\onefourthfigwidth}
\centering
\includegraphics[width=\textwidth]{figures/mcmc/kidney_ratio_150.png}
\end{subfigure}
\begin{subfigure}[b]{\onefourthfigwidth}
\centering
\includegraphics[width=\textwidth]{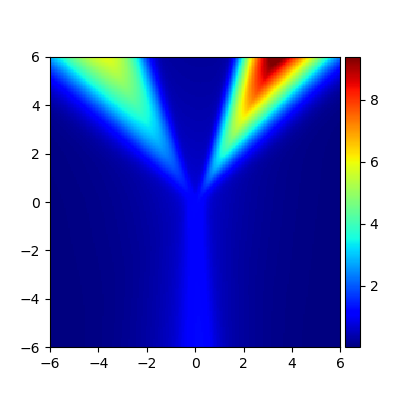}
\end{subfigure}
\begin{subfigure}[b]{\onefourthfigwidth}
\centering
\includegraphics[width=\textwidth]{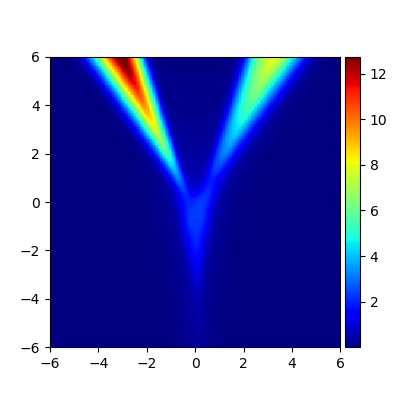}
\end{subfigure}
\begin{subfigure}[b]{\onefourthfigwidth}
\centering
\includegraphics[width=\textwidth]{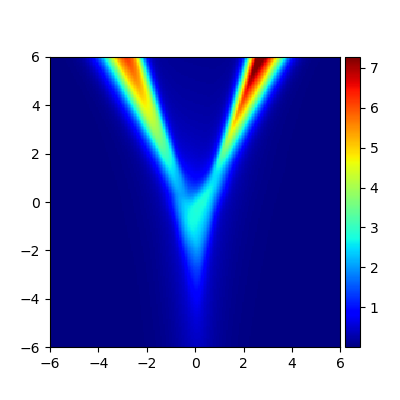}
\end{subfigure}
\begin{subfigure}[b]{\onefourthfigwidth}
\centering
\includegraphics[width=\textwidth]{figures/mcmc/banana_ratio_150.png}
\end{subfigure}
\caption{The~\algshort estimates after $\cbr{10, 20, 30, 150}$ iterations on the datasets. As we can see, with the algorithm proceeds, the learned stationary density ratio is getting closer to the ground-truth.}
\label{fig:mcmc_evolve}
\end{figure}

To verify the convergent behavior of our method, \cref{fig:mcmc_evolve} shows how the ratio network improves as we train the model. 
It can be seen that the our method quickly concentrates its mass to the region with high potentials. 

\subsection{Off-policy Evaluation}
\textbf{Taxi} is a $5\times5$ gridworld in which the taxi agent navigates to pick up and drop off passengers in specific locations.
It has a total of $2000$ states and $6$ actions. 
Each step incurs a $-1$ reward unless the agent picks up or drops off a passenger in the correct locations. 
The behavior policy is set to be the policy after $950$ Q-learning iterations and the target policy is the policy after $1000$ iterations. 
In the Taxi experiment, given a transition $(s,a,s')$, instead of sampling one single action from the target policy $\pi(a'|s')$, we use the whole distribution $\pi(\cdot|s')$ for estimation. 
We conduct closed-form update in the power method and the number of steps is $T=100$. 

\textbf{Continuous experiments}. 
The environments are using the open-source PyBullet engine.
The state spaces are in $\RR^9,\RR^{26},\RR^{28}$ respectively and the action spaces are in $\RR^2,\RR^6,\RR^8$ respectively.
the $\tau$ model is the same as in the SDE experiment (except for input, which depends on the environment). $T=200,M=10,B=1k$ and the learning rate is 0.0003. 
The model-based method has a similar neural network structure and is trained for $MT=2k$ steps with a learning rate of 0.0005. The target policy for the Reacher agent is pretrained using PPO while the HalfCheetah and Ant agents are pretrained using A2C (all with two hidden layers of $64$ units each). 

The results in the plots are mean and standard deviation from 10 runs.

\end{appendix}

\end{document}